\documentclass[runningheads]{llncs}
\usepackage[T1]{fontenc}
\usepackage{graphicx, floatrow}
\usepackage{booktabs} 
\usepackage{amsmath,url,graphicx,amssymb}
\usepackage{amsfonts}
\usepackage{multirow}
\usepackage{algorithm}
\usepackage{algpseudocode}
\usepackage{pifont}
\usepackage{color}
\usepackage{xspace}
\usepackage{enumitem}
\usepackage{subcaption}
\usepackage{colortbl}
\usepackage{xcolor}
\usepackage{bm}
\usepackage{caption} 
\usepackage{bbm}
\usepackage{todonotes}
\usepackage[
  separate-uncertainty = true,
  multi-part-units = repeat
]{siunitx}

\newcommand{\mylistbegin}{
  \begin{list}{$\bullet$}
   {
     \setlength{\itemsep}{-2pt}
     \setlength{\leftmargin}{1em}
     \setlength{\labelwidth}{1em}
     \setlength{\labelsep}{0.5em} } }
\newcommand{\mylistend}{
   \end{list}  }

\newcommand{\eg}{\textit{e.g.}}

\newcommand{\ie}{\textit{i.e.}}

\newcommand{\mR}{\mathcal{R}}
\newcommand{\mE}{\mathbb{E}}
\newcommand{\mX}{\mathcal{X}}
\newcommand{\mL}{\mathcal{L}}
\newcommand{\mS}{\mathcal{S}}
\newcommand{\mT}{\mathcal{T}}

\definecolor{sblue}{HTML}{02BCD4}
\definecolor{sred}{HTML}{F44436}
\definecolor{spink}{HTML}{E91E62}
\definecolor{sgreen}{HTML}{8BC34A}
\definecolor{spurple}{HTML}{3F51B5}
\definecolor{slightgreen}{HTML}{CCDE3A}
\definecolor{sorange}{HTML}{FE9800}
\definecolor{sgolden}{HTML}{FFC108}

\usepackage{amsmath}
\usepackage{amssymb}
\usepackage{dsfont}
\usepackage{graphicx}
\usepackage{xspace}
\usepackage{hyperref}
\begin{document}
\newcommand{\ours}{\texttt{Class Impression}\xspace}
\newcommand{\xiaoxiao}[1]{{\color{purple}[XL: #1]}}
\title{Class Impression for Data-free Incremental Learning}

%
%
\author{Sana Ayromlou\inst{1}
\and
Purang Abolmaesumi\inst{1}
\and
Teresa Tsang\inst{2}
\and
Xiaoxiao Li\inst{1}
}
\authorrunning{Ayromlou et al.}
\institute{
The University of British Columbia, Vancouver, BC, Canada
\and 
Vancouver General Hospital, Vancouver, BC, Canada\\
    \email{\{s.ayromlou,xiaoxiao\}@ece.ubc.ca}}

\maketitle              
\begin{abstract}
Standard deep learning-based classification approaches require collecting all samples from all classes in advance and are trained offline. This paradigm may not be practical in real-world clinical applications, where new classes are incrementally introduced through the addition of new data. Class incremental learning is a strategy allowing learning from such data. However, a major challenge is catastrophic forgetting, i.e., performance degradation on previous classes when adapting a trained model to new data. To alleviate this challenge, prior methodologies save a portion of training data that require perpetual storage, which may introduce privacy issues. Here, we propose a novel data-free class incremental learning framework that first synthesizes data from the model trained on previous classes to generate a \ours. Subsequently, it updates the model by combining the synthesized data with new class data. Furthermore, we incorporate a cosine normalized Cross-entropy loss to mitigate the adverse effects of the imbalance, a margin loss to increase separation among previous classes and new ones, and an intra-domain contrastive loss to generalize the model trained on the synthesized data to real data. We compare our proposed framework with state-of-the-art methods in class incremental learning, where we demonstrate improvement in accuracy for the classification of 11,062 echocardiography cine series of patients. Code is available at \href{https://github.com/sanaAyrml/Class-Impresion-for-Data-free-Incremental-Learning.git}{https://github.com/sanaAyrml/Class-Impresion-for-Data-free-Incremental-Learning}

\end{abstract}
\section{Introduction}

Deep learning classification models for medical imaging tasks have shown promising performance. Most of these models usually require collecting all training data and defining the tasks at the beginning. 
However, while highly desirable, it is impractical to train a deep learning model only once during deployment and then expect it to perform well on all future data, which may not be well represented in the training data. 
One promising solution is to allow the system to perform class incremental learning (a subset of continual learning or lifelong learning), \ie, adapting the deployed model to the newly collected data from new classes. 
Unfortunately, catastrophic forgetting~\cite{delange2021continual} occurs when a deep learning model overwrites past knowledge when training on new data.

Recent efforts to address catastrophic forgetting mainly include the following three categories: 
1) Replay methods, which alleviate forgetting by replaying some stored samples from previous tasks \cite{rebuffi2017icarl,rolnick2019experience,isele2018selective,chaudhry2019continual,hou2019learning}; 
2) Regularization-based methods, which use defined regularization terms such as additional losses to preserve prior learned knowledge while updating on new tasks \cite{li2017learning,jung2016less,rannen2017encoder,zhang2020class};
3) Parameter isolation methods, which allocate a fixed part of a static architecture for each task and only update that part during training \cite{serra2018overcoming,aljundi2017expert}.
These techniques have been mainly deployed in the natural image domain, and the solution for medical image analysis is under-explored~\cite{yang2021continual}. Among these approaches, rehearsal-based strategy has been reported to achieve the best results~\cite{hou2019learning}. Still, it requires accessing either the entire data or a portion of data representations used for training the previous model by saving them with more complex memory systems. It then performs re-training on the saved data with new data. This approach is less practical in medical imaging applications due to privacy regulations of data storage. 

To complement storing past data, data-free rehearsal (pseudo-rehearsal) strategies are recently proposed, in which the external memory is replaced with a model that is capable of generating samples from the past~\cite{smith2021always,lavda2018continual,shin2017continual}. 
However, the existing data-free rehearsal methods in class-incremental learning perform poorly for medical images (seen from our experiment results in Sec.~\ref{sec:exp}) without considering their unique properties. We reveal the \textbf{challenges} are: 1) generating synthetic images with high fidelity and preserving class-specific information from medical images is challenging; 2) mitigating the domain shift between generated and real images may not be straightforward in medical image classification; and 3) creating a robust decision boundary under imbalanced class distributions may be subject to significant uncertainty.  


It is practical yet challenging to deploy class-incremental learning for medical image analysis. To the best of our knowledge, no existing data-free rehearsal-based class incremental learning work on \textit{deep neural networks} is specifically designed for medical image analysis. Motivated by the observation that medical imaging samples within a categorical classification task normally share similar anatomical landmarks,
we propose a novel class-incremental learning pipeline that 1) restores \ours, \ie, generating prototypical synthetic images with high quality for each class using the frozen weights of the existing model, 2) mitigates the domain shift between class-wise synthesized images and original images by defining the \textit{intra-domain contrastive loss}, which empowers \ours in addressing the catastrophic forgetting problem, and 3) leverages a novel \textit{cosine normalized cross-entropy loss} for imbalance issue and a \textit{margin loss} to encourage robust decision boundary to regularize \ours for handling catastrophic forgetting and encouraging better generalization. We conduct extensive comparison experiments and ablation analysis on the echocardiogram view classification task to demonstrate the efficacy.

\section{Method}
\label{Methods}

\begin{figure}[t]
	\centering
	\includegraphics[width=1\textwidth]{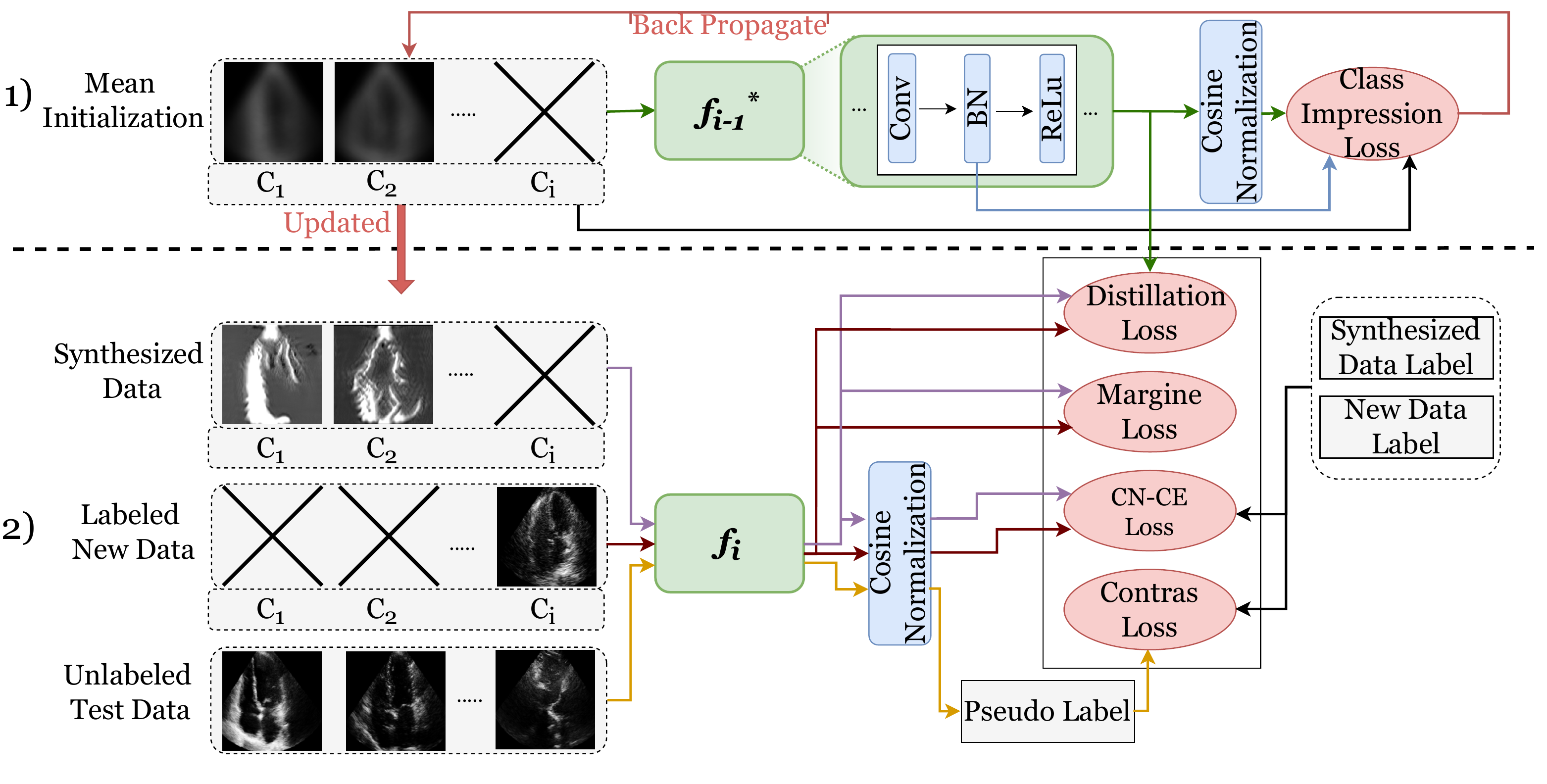}
		\caption{\ours consists of two main iterative steps to perform class incremental classification: 1) Synthesize samples images for each past class from the frozen model trained on the previous task $f_{i-1}^{*}$ by back-propagating using Eq.~\eqref{eq:CIloss} with class mean as initialization. 2) Update the model on new tasks. We utilize the distillation training method to leverage information from the previous model~\cite{hou2019learning,li2017learning}. We add Cosine Normalization cross-entropy loss~(Eq.~\eqref{eq:cosine}) and Margin loss over latent representation of synthesized images and labeled new data~(Eq.~\eqref{eq:margin}), to overcome data unbalance issue and catastrophic forgetting of past tasks, respectively. Furthermore, we introduce a novel Intra-domain Conservative loss~(Eq.~\eqref{eq:contras}) as a semi-supervised domain adaption technique between synthesized data and original data to mitigate the domain shift.}
	\label{thresh}
\end{figure}

\paragraph{Problem setting} The general class incremental learning considers a sequence of classification tasks, where in each task, new classes are added over time to a prior set of classes. The objective is to maintain high classification accuracy across all classification tasks. Let us denote $x^t \in \mathcal{X}^t$ as the input image, $y^t\in \mathcal{Y}^t$ as the class label w.r.t. the task $t$, and we have $\{\mathcal{Y}^t\} \subset \{\mathcal{Y}^{t+1}\}$. In rehearsal-based class incremental learning, the generalization error of all seen tasks is
\begin{align*}
    \sum_{t=1}^{T} \mathbb{E}_{(x^t,y^t)\sim(\mathcal{X}^t,\mathcal{Y}^t)}\left[ \ell(f_t(x^t;\theta), y^t)\right],
\end{align*}
with loss function $\ell$, classifier parameter $\theta$, the current task $T$, and the model $f_{t}$ trained in task $t$. While under data-free setting, there is no access to data $(\mathcal{X}^t,\mathcal{Y}^t)$ for $t<T$. 



\paragraph{Overview of our pipeline} Fig.~\ref{thresh} shows the whole pipeline of our proposed method. We have two stages for each task. In the first stage, we synthesize images from values saved in the frozen model trained on the previous task to employ them as replay data by seizing a genuine class-wise impression instead of saving them. In the second stage, we perform distilled class incremental training on the data using the new classes and synthesized data with proposed novel losses to 1) learn features of new classes, 2) preserve features comprehended from prior classes during previous tasks, and 3) maximize distances between the distribution of the new and previous classes in the latent space. Next, we will introduce each innovative component in our pipeline. 

\subsection{Class Impression}
Different from \cite{smith2021always} that trains a model to synthesize images without considering preserving class-specific information, we are motivated by recent work, DeepInversion~\cite{yin2020dreaming}, which was originally formulated to distill knowledge from a trained model for transfer learning. Given a randomly initialized input $\hat{x} \in  \mathbb{R}^{B\times H \times W \times C}$ where $B,H,W,C$ are the batch size, height, width, and number of channels, respectively, a target label $y$, the image is synthesized by optimizing
\begin{align}
\label{eq:CIloss}
    \min_{\hat{x}}\mathcal{L}_{\rm CE}(\hat{x},y)+\mR(\hat{x}),
\end{align}
where $\mathcal{L}_{\rm CE}(\cdot)$ is cross-entropy loss, and $\mR(\cdot)$ is the regularization to improve the fidelity of the synthetic images. Specifically, following DeepInversion~\cite{yin2020dreaming} and DeepDream~\cite{mordvintsev2015inceptionism}, we have
\begin{align}
    \mR(\hat{x})=\alpha_{\rm tv}\mR_{\rm TV}(\hat{x}) + \alpha_{\ell_2}\mR_{\ell_2}(\hat{x}) + \alpha_{\rm bn}\mR_{\rm BN}(\hat{x},\mathcal{X}),
\end{align}
where $\alpha$s are scaling factors, $\mR_{\rm TV}(\cdot)$ and  $\mR_{\ell_2}(\cdot)$ penalize the total variance and the $\ell_2$ norm of the generated image batch $\hat{x}$, respectively. $\mR_{\rm BN}(\cdot,\cdot)$ is for matching the  batch normalization (BN) statistic in trained model with the original images and defined as 
\begin{align}
    \mR_{\rm BN} = \sum_l \|\mu_l(\hat{x} - \mE(\mu_l(x)|\mX)\|_2 + \sum_l \|\sigma_{l}^2(\hat{x} - \mE(\sigma_{l}^2(x)|\mX)\|_2,
\end{align}
where $\mu_l$ and $\sigma_{l}^2$ are the batch-wise mean and variance estimates of feature maps corresponding to the $l$-th BN layer. To inject class-specific impression to the synthetic images without violating privacy regulations (such as re-identification), we assume the averaged image of each class $\bar{x}_k^t$ ($k\in \mathcal{Y}^t$) is available to initialize the optimization of $\hat{x}$ in Eq~(\ref{eq:CIloss}).  It is worth noting that, \ours aims to \textit{generate} images following the distributions of the past class, rather than reconstructing the training data points as model inversion attacks does~\cite{yin2021see,huang2021evaluating,hatamizadeh2022gradient}, thus \ours aims to meet certain privacy requirements imposed on storing medical data.


\subsection{Novel Losses}
Although we can leverage \ours, a.k.a the synthetic data of the past classes to conduct rehearsal distillation-based training, which mitigates catastrophic forgetting, class weights of old classes still may be ill-updated and mismatched with the updated representation space~\cite{zhu2021class}. Therefore, we propose to use a modified classification loss, \ie, \textit{cosine normalized cross-entropy loss}, and two additional regularization terms, \ie, \textit{margin loss}, and \textit{intra-domain contrastive loss} to improve the utility of \ours.
\subsubsection{Cosine Normalized Cross-entropy Loss}
Classifier bias commonly exists in class incremental learning, because data in the new classes are more abundantly available or mostly have better quality~\cite{zhu2021class}. In classification task, the Softmax operation yields the prediction probably of class $k$ as $p_i(x) = \exp(\theta_i^{\top}f(x)+b_k)/\sum_j\exp(\theta_j^{\top}f(x)+b_i)$, which $f(\cdot)$ is feature extractor, and $\theta_k$ and $b_k$ are class embedding and bias weights. Due to class imbalance, the magnitudes of embedding and bias of the new class can be significantly higher than the past classes. Cosine normalization has been widely used in vision tasks~\cite{gidaris2018dynamic,hou2019learning,qi2018low} to eliminate the bias caused by the significant difference in magnitudes. Thus, following the above-mentioned work \cite{hou2019learning}, we add the following \textit{Cosine Normalized Cross-entropy} (CNCE) loss for classification with the past and new classes at the current task $t$:
\begin{align}
\label{eq:cosine}
    \mL_{\rm CNCE} = -\sum_{k=1}^{|\mathcal{Y}^t|}y_k\log\left(\frac{\exp(\eta\langle \bar{\theta}_k,\bar{f}(x)\rangle )}{\sum_j\exp(\eta\langle \bar{\theta}_j,\bar{f}(x)\rangle )}\right),
\end{align}
where $\bar{v}=v/\|v\|_2$ indicates the unit normalized vector,  $\langle\cdot,\cdot\rangle$ measures the cosine similarity of two vectors, and $\eta$ is the temperature hyperparameter. Further, \ours can generate an equal number of data for each past class as that of the new class to alleviate the imbalanced issue. 
\subsubsection{Margin Loss}
Representation overlapping is another inherent problem in class incremental learning~\cite{zhu2021class}. Margin loss uses a margin to compare samples representations distances, and is used in few-shot learning~\cite{li2020boosting} and non-data-free class incremental learning~\cite{hou2019learning}. To avoid the ambiguities between the past and new classes, we are the first to explore margin loss in data-free class incremental learning that encourage separating the decision boundary of the new class from the old ones. At the current task $t$, margin loss is written as
\begin{align}
\label{eq:margin}
    \mL_{\rm margin} = \sum_{k=1}^{|\mathcal{Y}^{t-1}|}\max \left( m - \langle \bar{\theta} , \bar{f}(x) \rangle + \langle \bar{\theta}^k,\bar{f}(x)\rangle, 0 \right),
\end{align}
where $x$ are the synthesized images that are used as anchors of the classes seen in the previous tasks' distribution, $m$ is the margin for tolerance, $\bar{\theta}$ is the embedding vector of $x$'s true class, and $\bar{\theta}^k$ is the embedding vector of the new class, which is viewed as negatives for $x$. A larger $m$ encourages a larger separation.

\subsubsection{Intra-domain Contrastive Loss}
We notice that directly applying the synthetic medical images for rehearsal may not generalize well to classify the original data. Motivated by semi-supervised domain adaption~\cite{singh2021clda}, we aim to align the domain shift between the source $\mS$ (\ie, synthetic images) and target $\mT$ (\ie, unlabeled testing data in the new task). The centroid of the images from the source domain belonging to class $k$ is defined as $c_k^\mS = \sum_i\mathbbm{1}_{\{y_i^{\mS}=k\}}f(x_i^s)/\sum_i\mathbbm{1}_{\{y_i^{\mS}=k\}}$. Then, we assign the pseudo label for the unlabeled target samples using the currently updated classifier. An intra-domain contrastive loss can pull the source and target samples in the same class together and push apart the centroids of different classes, which is written as (at task $t$):
\begin{align}
\label{eq:contras}
    \mL_{\rm contras} (c_k^\mS,c_k^\mT) = -\log \frac{\exp(\tau\langle \bar{c}_k^\mS, \bar{c}_k^\mT \rangle)}{\exp(\tau\langle \bar{c}_k^\mS, \bar{c}_k^\mT \rangle) + \sum_{\substack{j=1 \\ \mathcal{Q}\in\{\mS,\mT\}}}^{|\mathcal{Y}^{t-1}|}\mathbbm{1}_{\{j\neq k\}}\exp(\tau\langle \bar{c}_k^\mathcal{Q}, \bar{c}_k^\mT \rangle) },
\end{align}
where $\tau$ is the temperature hyperparameter in $\exp(\cdot)$.

Altogether, the total loss used to train the new model for recognizing the new class while maintaining the old knowledge is given by:
\begin{align}
\label{eq:total}
\mL_{\ total} = \mL_{\rm CNCE} + \alpha_{\rm dist} \mL_{\rm dist} + \alpha_{\rm margin}\mL_{\rm margin} + \alpha_{\rm contras}\mL_{\rm contras},
\end{align}
where $\alpha$ are tunable scaling factors, and $\mL_{\rm dist} = 1 - \langle \bar{f}^*(x), \bar{f}(x) \rangle$ with $\bar{f}^*$ as the old model, is the distillation loss widely used in class incremental learning~\cite{hou2019learning,li2017learning}.

\section{Experiments}
\label{sec:exp}
\subsection{Datasets and Experimental Settings}

\noindent \textbf{Dataset:} Heart echo data utilized in this paper come from several investigations at our local institution, randomly chosen from the hospital picture archiving system, with authorization from the Information Privacy Office and Clinical Medical Research Ethics Board. These cines are captured by six devices: GE Vivid 7, Vivid i, Vivid E9, Philips iE33, Sonosite, and Sequoia. Our dataset contains 11,062 cines (videos) of 2151 unique patients diagnosed with various heart diseases, with an average of 48 frames in each. An experienced cardiologist labeled gathered cines as five different views. The view distribution is shown in Supplementary Table 1. We chose our task as a test-bed to evaluate our innovations. In contrast to many other medical imaging classification problems, the labels associated with standard echo views are less ambiguous, less noisy, and anatomically interpretable, making the analysis of classification results easier and its failure modes tractable. 
Finally, we split the data into training, validation, and test sets based on the subject with the ratio of 70\%, 20\%, and 10\%, respectively.
\\

\noindent \textbf{Experimental Settings}: We perform a four-task class incremental learning, with a two-way classification at the beginning and adding in one new class at each task. As the number of data points is relatively low compared to natural image's benchmark datasets, the classification model of \ours is a ResNet-based convolutional neural network with three residual blocks, each containing two layers. All models are implemented with PyTorch and trained on one NVIDIA Tesla V100 GPU with 16GB of memory.

As mentioned in Section \ref{Methods}, our implementation contains two main stages for each task. We \textbf{freeze} model weights for the image synthesis stage. We use Adam optimizer and optimize the batch of 40 images with a learning rate of 0.01 for 2000 epochs. 
In the stage of incremental learning with a new class, we \textbf{update} model weights again with the SGD optimizer and continue to learn on each new task for 30 epochs with a batch size of 40. We search the optimal hyper-parameters 
using Weights \& Biases Sweeps, an automated hyperparameter searching software~\cite{wandb}. The best chosen hyperparameters are detailed in Supplementary.

\begin{figure}[t]
\floatsetup{valign=t, heightadjust=all}
	\centering
        	

	\begin{subfigure}[t]{\textwidth}
    	\centering
         \begin{subfigure}[t]{0.485\textwidth}
        	\includegraphics[width=1\textwidth]{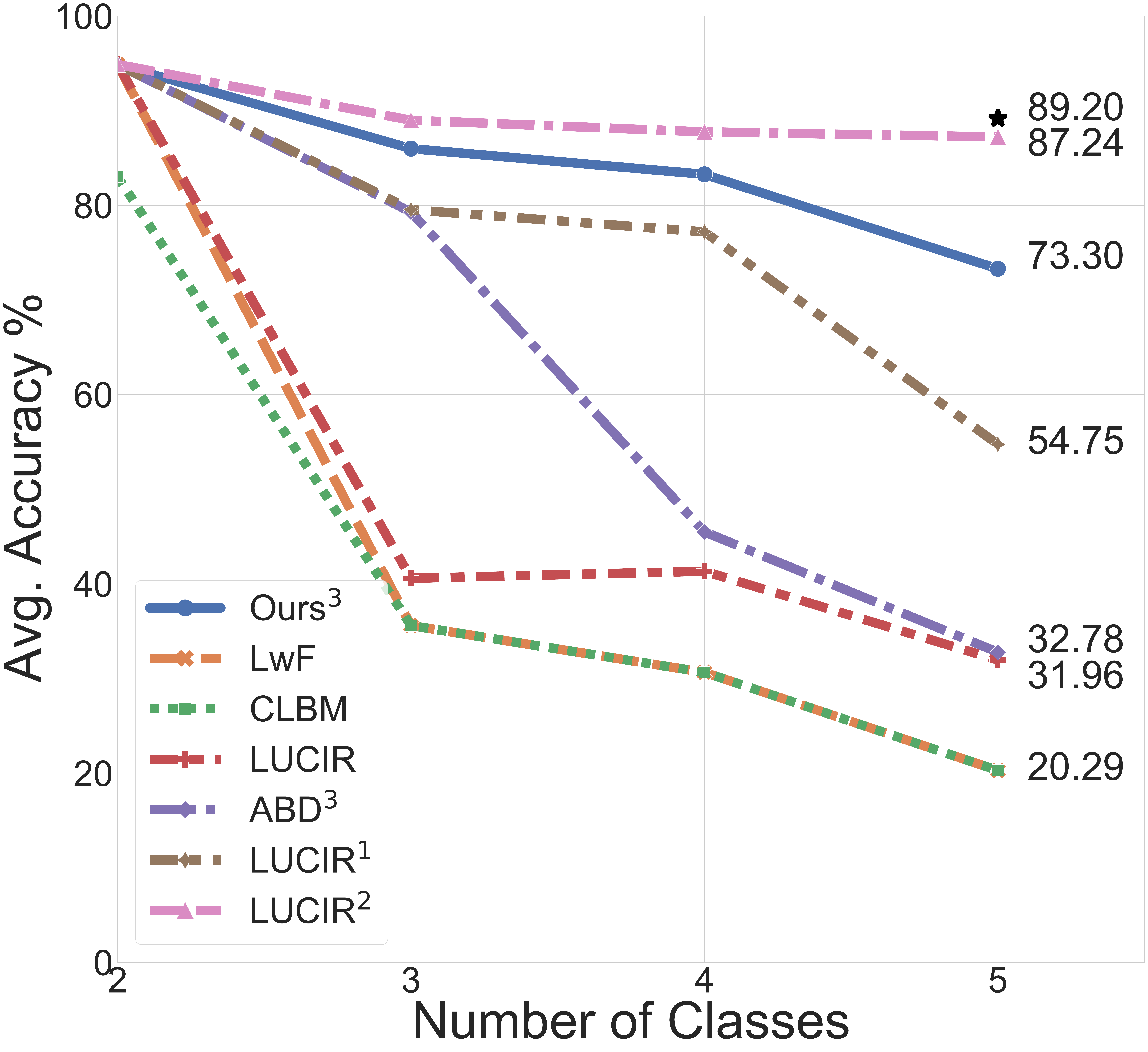}
        	\caption{Comparision with Baselines \footnote{ 
        	1. With corset data; 2. With weighted corset data; 3. With synthesized data }}
        	\label{fig:baseline}
        \end{subfigure}
        \hfill
        \begin{subfigure}[t]{0.485\textwidth}
        	\includegraphics[width=1\textwidth]{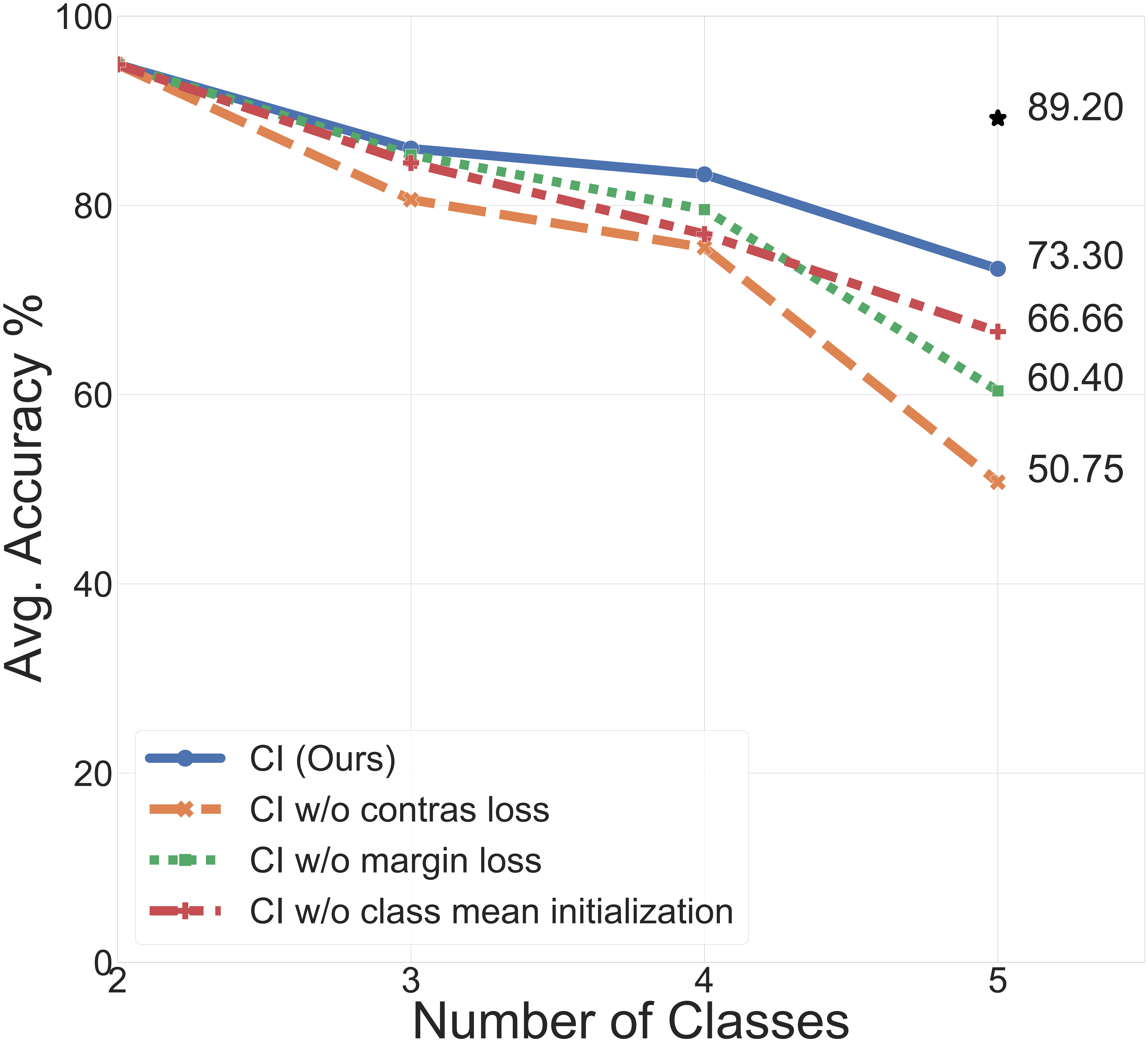}
        	 \caption{Ablation Study}
            \label{fig:abalation}
        \end{subfigure}
        \hfill
    	
	\end{subfigure}
	\caption{ Testing accuracies of four tasks. It shows the classification accuracies on the heart echo test dataset comparing with (a) baselines of class-incremental learning and (b) ablation study on different configurations of proposed framework. $\star$ indicates the accuracy of Oracle model, \ie, training all the classes offline w/o class-incremental setting. \ours shows consistently higher accuracy with an average of 31.34\% increase after the final task compared to the SOTA data-free incremental learning (ABD).
}
	\label{compare_table}
\end{figure}
\subsection{Comparison with Baselines}

We compare \ours with the best-performed data-free baselines and a non-data-free baseline that saves coreset data in the incremental learning setting by implementing them on our heart echo dataset. 

The baseline methods included in this study are listed as follows:
\textbf{LwF}\cite{li2017learning} uses distillation-based training to overcome catastrophic forgetting among different tasks. \textbf{ABD}\cite{smith2021always} generates synthesized images of classes in the prior task. \textbf{CLBM}\cite{yang2021continual} fits Gaussian mixture models to save information extracted from data points in each class. \textbf{LUCIR}\cite{hou2019learning} is a state-of-the-art rehearsal-based method that saves a coreset of data points from each seen class and replays them as anchors of their respective class distribution to maximize the distance between the distribution of classes in the latent space.

The results are given in Fig.~\ref{fig:baseline}. \textbf{LWF}, reaching $20.29\%$ accuracy, completely fails to consider the imbalance issue among classes in different tasks and distinguish between inter-task classes. \textbf{ABD} is a recent work that shares a similar data synthesis idea with us for class incremental learning. However, its final task accuracy is only $32.78\%$, much worse than \ours. Note that our innovations over \textbf{ABD} are multifold, lying in utilizing the unique class-specific information in medical images: 1) we initialize batches for the image synthesis step with the mean of each class; and 2) we mitigate the domain shift between class-wise synthesized images and original images by defining the \textit{intra-domain contrastive loss}. 
\textbf{CLBM} overfits to new data feature distribution and fails to distinguish it from the old ones, yielding poor classification accuracy. \textbf{LUCIR} saves auxiliary information (\ie, coreset data of old classes) and performs the closest to \ours. If we implement it without storing any data or saving a determined number of data as coreset from each class, it reaches the accuracy of $54.75\%$ and \ours fairly outperforms it. However, if we use a weighted sampler in \textbf{LUCIR} to increase the saved samples' effect, it exceeds \ours. This result is expected due to intrinsic information loss in the data-free setting. In comparison, \ours results substantially outperform prior class incremental data-free settings by boosting $31.34\%$ accuracy on the all seen classes in the final task and standing on $73.3\%$ accuracy over five classes. The synthetic images generated by \ours are presented in Supplementary.



\subsection{Ablation Studies}


We perform extensive ablation studies to expose the impact of different losses in the final performance. The results are given in Fig.~\ref{fig:abalation}.
1) \textit{Impact of domain adaption:} We omit intra-domain contrastive loss~(Eq.~\eqref{eq:cosine}) to observe the effect of the gap between the domain of synthesized images and original images. As seen in Fig.~\ref{fig:abalation}, it results in the worst performance compared to other settings.
2) \textit{Impact of inter-class separation:} When we omit margin loss~(Eq.~\eqref{eq:margin}), the model does not increase the distance of data points from previous tasks and new tasks, which also leads to a significant accuracy drop.
3) \textit{Impact of mean initialization:} We initialize input images with random Gaussian noise instead of the mean of each class in the synthesis stage. Therefore, the model fails to capture the true impression of each class, and performance reduces. 

\section{Conclusion}
In this work, we propose \ours, a novel data-free class incremental learning framework. In \ours, instead of saving data from classes in the earlier tasks that are not available for training in the new task, we synthesize class-specific images from the frozen model trained on the last task.
Following, we continue training on new classes and synthesized images using the proposed novel losses to alleviate catastrophic forgetting, imbalanced data issues among new and past classes, and domain shift between new synthesized and original images of old classes. 
Experimental results for echocardiography cines classification on the large-scale dataset validate \ours out-performs the SOTA methods in data-free class incremental learning with an improbable gap of 31.34\% accuracy in the final task and get comparable results with the SOTA data-saving rehearsal-based methods.

Our proposed method shows the potential to apply incremental learning in many healthcare applications that cannot save data due to memory constraints or privacy issues. It is common in clinical deployment that a client inherits a trained model without having access to its training data. Our design enables the client to refine the model with new tasks. For our future work, we plan to combine our work with other practical settings in the real world in the medical domain (\eg, testing on different medical datasets) and further improve the pipeline to meet clinical-preferred performance.
 
\section*{Acknowledgement}
This work is supported in part by the Natural Sciences and Engineering Research Council of Canada (NSERC), the Canadian Institutes of Health Research (CIHR), and NVIDIA Hardware Award. We thank Dr. Hongxu Yin at NVIDIA research for his insightful suggestions.

%
%
%
\bibliographystyle{splncs04}
\bibliography{reference.bib}
%




\end{document}